# An efficient real-time target tracking algorithm using adaptive feature fusion

**Yanyan Liu[a], Changcheng Pan[a,c], Minglin Bie[a], and Jin Li[b*]**

[a]Department of Electronics and Information Engineering, Changchun University of Science and Technology, Jilin 130022, China
[a]Departamento de Óptica, Facultad de Física, Universidad Complutense, 28040 Madrid, Spain
[c]Zhejiang Saisi Electronic Technology Co., Ltd., Jiaxing Science and Technology City, Zhe

*Corresponding author: Jin Li (E-mail: jinli02@ ucm.es).

**ABSTRACT** Visual-based target tracking is easily influenced by multiple factors, such as background clutter, targets' fast-moving, illumination variation, object shape change, occlusion, etc. These factors influence the tracking accuracy of a target tracking task. To address this issue, an efficient real-time target tracking method based on a low-dimension adaptive feature fusion is proposed to allow us the simultaneous implementation of the high-accuracy and real-time target tracking. First, the adaptive fusion of a histogram of oriented gradient (HOG) feature and color feature is utilized to improve the tracking accuracy. Second, a convolution dimension reduction method applies to the fusion between the HOG feature and color feature to reduce the over-fitting caused by their high-dimension fusions. Third, an average correlation energy estimation method is used to extract the relative confidence adaptive coefficients to ensure tracking accuracy. We experimentally confirm the proposed method on an OTB100 data set. Compared with nine popular target tracking algorithms, the proposed algorithm gains the highest tracking accuracy and success tracking rate. Compared with the traditional Sum of Template and Pixel-wise LEarners (STAPLE) algorithm, the proposed algorithm can obtain a higher success rate and accuracy, improving by 2.3% and 1.9%, respectively. The experimental results also demonstrate that the proposed algorithm can reach the real-time target tracking with 50+fps. The proposed method paves a more promising way for real-time target tracking tasks under a complex environment, such as appearance deformation, illumination change, motion blur, background, similarity, scale change, and occlusion.

## I. INTRODUCTION

Target tracking using visual information is a very important tool in many fields, such as civil and military [1], automatic driving, intelligent monitoring, etc. The accuracy of tracking algorithms can directly determine the success of a tracking task. Unfortunately, current tracking algorithms have relatively low accuracy in a complex imaging environment because the tracked target's visual information is easily disturbed by many factors, such as background clutter, targets' fast-moving, illumination variation [2-4], object shape change, occlusion, etc.

To address this problem, many target tracking algorithms [5]-[7] have been proposed. Currently, target tracking algorithms widely adopt a discriminative model method, such as minimizing the output sum of squared error (MOSSE) [8], exploiting the circulant structure of tracking-by-detection with Kernels (CSK) [9], Kernelized Correlation Filter (KCF) [10-11], STAPLE, Discriminative Scale Space Tracker (DSST) [12]-[13]. Current investigations demonstrated that the histogram of oriented gradient (HOG) feature [14, 15] is a valuable parameter in the target tracking process, which was widely used in target tracking algorithms. The HOG feature is good for motion blur and light change. Unfortunately, the HOG feature is not good for deformation and fast motion.

To improve the target tracking accuracy, a multiple-feature-based fusion method is the necessary step of target tracking algorithms. A kernel-related tracking method based on adaptive feature fusion was proposed to reduce illumination sensitivity (IS) [16]. The STAPLE algorithm combines the histogram of oriented gradient (HOG) [14] and color statistical features to find the target location. As mentioned in HOG, the HOG feature is good for motion blur and light change. The color feature is not sensitive to deformation, and it is not good for light changes and background color similarity. The two feature fusion can compensate for the corresponding drawbacks. Through these two feature fusions, the STAPLE algorithm has a very good tracking accuracy [17-24]. However, the conventional STAPLE algorithm couldn't get the optimal fusion results because it adopts constant coefficients to perform the two-feature fusion.

In this paper, we propose a real-time target tracking method based on a low-dimension adaptive feature fusion capable of the simultaneous implementation of high-accuracy and real-time target tracking. In the STAPLE algorithm, we found that the updated model includes too many updated parameters at every updating operation, which indicates the updated model needs to process the high dimension data. The high dimension process results in the slow speed of tracking algorithms and easily produces an over-fitting issue. Moreover, the HOG and color feature fusion is based on a constant-coefficient rule, which limits the tracking accuracy under different circumstances. The proposed method adopts a low-dimension adaptive feature fusion algorithm composed of three aspects. First, the adaptive fusion of the HOG feature and color feature is performed to improve tracking accuracy. Second, the proposed method adopts a convolutional dimension reduction method to the fusion between the HOG feature and color feature to reduce the over-fitting problem caused by their high dimensions fusions. Third, an average correlation energy estimation method is used to extract the relative confidence adaptive coefficients to ensure tracking accuracy. In the average correlation energy estimation method the fusion coefficients can determine the best filter from a correlation filter and a Bayes classifier [25]-[33] to better complete the tracking. Based on these advantages, the proposed algorithm can implement an accurate target tracking task at a real-time condition.

## II. PROPOSED ALGORITHM

### A. STAPLE ALGORITHM

The HOG feature is a gradient feature of cell size. The HOG feature has a good tracking performance within an image local area. However, The HOG feature has an inferior performance within a whole image area. The color features are achieved by extracting the color histogram of a global image [34]-[41], which has a good tracking performance within a whole image area. The STAPLE algorithm can fully fusion the two features to perform the tracking algorithm.

The position of an observed target in the $t^{th}$ frame is defined by the maximum response (having the highest score) of a rectangular box (denoted by $p$) in the image $x_t$, in which the maximal rectangular box can be expressed as:

$$p_t = \arg\max_{p \in S_t} f(T(x_t, p); \theta_{t-1}), \tag{1}$$

where $T(\cdot)$ expresses an image transform function, $\theta$ is a model parameter, $S_t$ is the rectangular box set in the $t$-th frame. In Eq.1, using different model parameters $\theta$, $f(T(x_t, p); \theta_{t-1})$ can generate different scores of the rectangular box $p$. To obtain the highest score, the optimum model parameters should ensure minimizing a loss function. The loss function is denoted by $L(\theta; X_t)$, where $X_t$ is an image set consisting of the previous images and the location of the observed target. That is $X_t=(x_i, p_i)_i^t$. In Eq.1, the $t$-th frame model parameter $\theta_t$ can be obtained by minimizing a loss function as

$$\theta_t = \arg\max_{\theta \in Q} \{L(\theta; X_t) + \lambda R(\theta)\}, \tag{2}$$

where $Q$ is the model parameter space, $R(\theta)$ is a regularization term, and $\lambda$ is the weight. To efficiently and fast solve Eq.1 and Eq.2, $f$ is realized by a linear combination of a template and a histogram as

$$f(x) = \gamma_{tmpl} f_{tmpl}(x) + \gamma_{hist} f_{hist}(x), \tag{3}$$

where $\gamma_{tmpl}$ and $\gamma_{hist}$ are the weight of the template and histogram score allocations, $f_{tmpl}$ is a score allocation function of the template, and $f_{hist}$ is a score allocation function of the histogram. $f_{tmpl}$ can be defined by a linear combination equation as

$$f_{tmpl}(x; h) = \sum_{\mu \in T} h[\mu]^T \phi_x[\mu], \tag{4}$$

where, $\phi_x$ is a featured image obtained from $x$, $h$ is the weight vector obtained from $x$, $T$ is a patch set in $x$, and $u$ represents a pixel position of a patch in $T$. Accordingly, $f_{hist}(x)$ of Eq.3 can be expressed as

$$f_{hist}(x; \beta) = g(\varphi_x; \beta) \tag{5}$$

with

$$g(\varphi_x; \beta) = \beta^T \left(\frac{1}{H} \sum_{\mu \in H} \varphi[\mu]\right), \tag{6}$$

where $\varphi_x$ is a featured image obtained from $x$, $\beta$ is the model parameter, and $H$ is a patch set in a feature image $\varphi_x$. Eq.4 is the convolution operation of HOG template $h$ and the patch of HOG features, which is equivalent to a CF filter. By calculating the HOG feature, a correlation filtering tracking algorithm is performed to obtain the fraction of $f_{tmpl}$. Through calculating the color features, the fraction of $f_{hist}$ is obtained. The combination with two obtained fractions is used to complete the fusion.

In Eq.2, to accelerate the fusion calculation, the overlapping windows should share a feature calculation. The histogram scores can be calculated by an integral image. We define the training loss using a weighted linear combination of single image loss as:

$$L(\theta; X_T) = \sum_{t=1}^{T} w_t l(x_t, p_t, \theta), \tag{7}$$

where $L(\theta, X_T)$ is the training loss function, $l(x_t, p_t, \theta)$ is the frame's loss function, and $w_t$ is coefficient. In Eq.7, the loss function of each image is:

$$l(x, p, \theta) = d\left(p, \arg\max_{p \in S} f(T(x, q); \theta)\right), \tag{8}$$

where $d(p, q)$ defines the cost of choosing rectangle $q$ when the correct rectangle is $p$.

In Eq.2, the parameter of the whole model is $\theta=(h, \beta)$. The speed and effectiveness of the correlation filter can be obtained by solving two independent ridge regression problems to learn the model as:

$$h_t = \arg\min_h \left\{ L_{tmpl}(h; X_t) + \frac{1}{2}\lambda_{tmpl} \|h\|^2 \right\}, \tag{9}$$

$$\beta_t = \arg\min_\beta \left\{ L_{hist}(\beta; X_t) + \frac{1}{2}\lambda_{hist} \|\beta\|^2 \right\}, \tag{10}$$

where $h_t$ and $\beta_t$ are two independent ridge regression.

When $L(\theta; x)$ is a convex function of $f(\theta; x)$ and $f(\theta; x)$ is linear with respect to $\theta$, there are a matrix $A_t$ and a vector $b_t$ to meet the following equation as:

$$L(\theta, X_t) + \lambda \|\theta\|^2 = \frac{1}{2}\theta^T (A_t + \lambda I)\theta + b_t^T \theta + const. \tag{11}$$

In the case of least-squares correlation filtering, the loss of each image is expressed as:

$$l_{tmpl}(x, p, h) = \left\| \sum_{k=1}^{K} h^k * \phi^k - y \right\|^2, \tag{12}$$

where $h$ is the input image, $\phi$ is is the filter template, $y$ is the ideal response, $l_{tmpl}(x,p,h)$ is the loss function, and $h^k$ refers to channel $k$ of multi-channel image $h$. Therefore the normalized objective function is:

$$\hat{h}[\mu] = (\hat{s}[\mu] + \lambda I)^{-1} \hat{r}[\mu], \tag{13}$$

where $\hat{s}[\mu]$ is a $K \times K$ matrix with elements and $\hat{r}[\mu]$ is a $K$-dimensional vector with elements $\hat{r}^i[\mu]$.

Accordingly, the histogram loss of each image is expressed by:

$$l_{hist}(x, p, \beta) = \sum_{(q,y)} \left( \beta^T \left[ \sum_{\mu \in H} \varphi_{T(x,q)}[\mu] \right] - y \right)^2. \tag{14}$$

The histogram score is obtained by the average vote in the feature RGB color. Here, the target function of each image is used for linear regression of each feature pixel as:

$$l_{hist}(x, p, \beta) = \frac{1}{O}\sum_{\mu \in o}(\beta^T \varphi[\mu] - 1)^2 + \frac{1}{B}\sum_{\mu \in B}(\beta^T \varphi[\mu])^2, \tag{15}$$

where $O$ and $B$ is the pixels of an image.

## B. CONVOLUTION DIMENSION REDUCTION

In this paper, we adopt an improved scheme on feature fusion, which is convenient for the algorithm to track targets better after extracting features from STAPLE. The features are transformed into the continuous spatial domain by an interpolation operation when the features are extracted as

$$J_d\{x^d\}(t) = \sum_{n=0}^{N_d - 1} x^d[n] b_d\left(t - \frac{T}{N_d}n\right). \tag{16}$$

In the continuous formulation, the operator $S_f$ is parameterized by a set of convolution filters $f = (f^1, \ldots, f^d, \ldots, f^D) \in L^2(T)^D$. Here, $f^d \in L^2(T)$ is the continuous filter for the feature channel $d$. Here, the * denotes the convolution operator. $J(x)$ is the extracted feature. The final detection score is calculated as:

$$S_f\{x\} = f * J(x) = \sum_{d=1}^{D} f^d * J_d\{x^d\} \quad , \tag{17}$$

where *f* is the characteristics of each dimension. The objective function of the learning correlation filter is converted to the frequency domain as:

$$E(f) = \sum_{j=1}^{M} \alpha_j \left\| S_f\{x_j\} - y_j \right\|_{L^2}^2 + \sum_{d=1}^{D} \left\| \omega f^d \right\|_{L^2}^2 \quad , \tag{18}$$

where $y_i$ is the desired output of the convolution operator, $S_f\{x_i\}$ is applied to the training sample $x_i$, and $\alpha_i$ controls the impact of each training sample.

In the feature extraction using Eq.17 and Eq.18, there are D filters corresponding to the extraction of D-dimensional features, but many of them have a small contribution. We select the C-dimension from the D-dimensional features. The new detection function is defined by $S_{pf}(x)$, where *p* is the matrix of $D \times C$. The new detection function can be expressed as:

$$S_{pf}\{x\} = pf * J(x) = \sum_{c,d} p_{d,c} f^c * J_d\{x^d\} = f * J\{x\} \quad , \tag{19}$$

$$E(f, p) = \left\| z^T Pf - y \right\|_{L^2}^2 + \sum_{c=1}^{C} \left\| w * f^c \right\|_{L^2}^2 + \lambda \left\| p \right\|_{L^2}^2 \quad . \tag{20}$$

All *C* filters are obtained by linear combination to form the features of a dimension to represent each row of the matrix. The target video sequence is studied in the first frame, and the subsequent video frame remains unchanged during the tracking process. The new objective function of the learning correlation filter is $z = J\{x\}$, where $z^T Pf$ is bilinear. A Gauss-newton iterative method and conjugate gradient operation are used from the *D* dimension to the *C* dimension. This new process can significantly improve the speed of the algorithm and prevent the phenomenon of over-fitting.

## C. ADAPTIVE FEATURE FUSION

In the STAPLE algorithm, the fusion of color features and HOG features is used to improve the tracking accuracy of a target task. However, the STAPLE algorithm adopts constant coefficients to achieve the fusion of these two features. The constant-coefficient-based fusion rule can be expressed as:

$$response = (1-\alpha) response\_cf + \alpha \times response\_p \quad , \tag{21}$$

where *response_cf* is the response of the kernel correlation filters, *response_p* is the response for a Bayesian classifier. The drawback of this processing using Eq.21 is that the target tracking results are unstable under different circumstances.

To address this issue, we proposed an adaptive feature fusion method. The principle of the proposed method is that we introduce an average peak correlation energy ratio (APCE) to the target tracking process to improve the target tracking accuracy in different circumstances. The APCE represents the fluctuation degree of a response graph and the confidence of the detection target [9, 42]. The APCE can be expressed as:

$$APCE = \frac{\left| F_{\max} - F_{\min} \right|^2}{mean\left( \sum \left( F_{w,h} - F_{\min} \right)^2 \right)} \quad , \tag{22}$$

where $F_{max}$ is the maximum value of the $F_{w,h}$ response and $F_{min}$ is the minimum value of the $F_{w,h}$ response. When a target (having a sharper peak and less noise) within the detection range is obvious, the APCE value becomes larger and the response graph becomes smoother with only one peak. Conversely, if a target is obscured or lost, the APCE is decreased significantly. Fig.1 shows the APCE value in these cases. When the target is severely deformed or blocked, the confidence response graph presents multi-peak irregularities, and the APCE of the response graph decreases rapidly from 14.9 to 3.6.

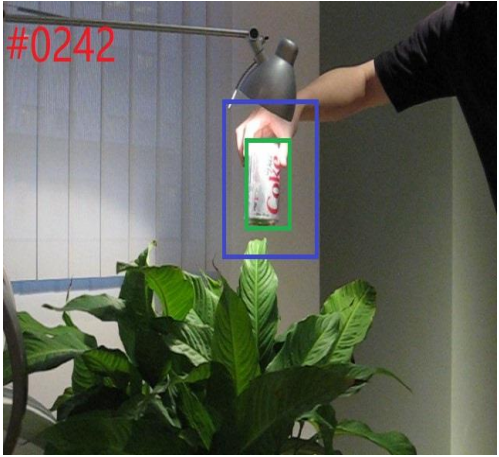

(a)

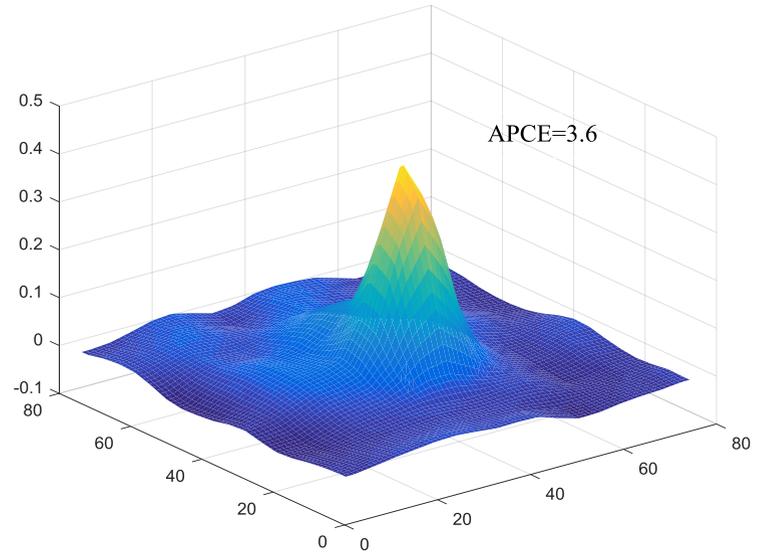

(b)

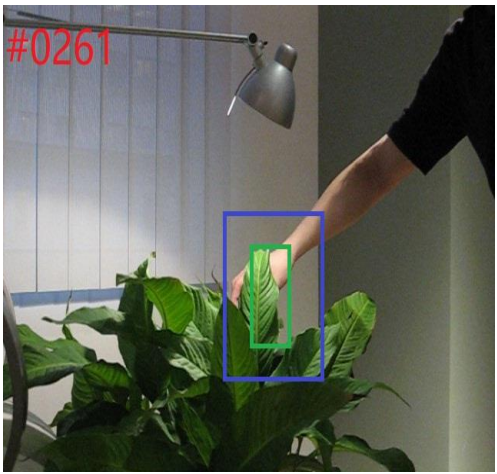

(c)

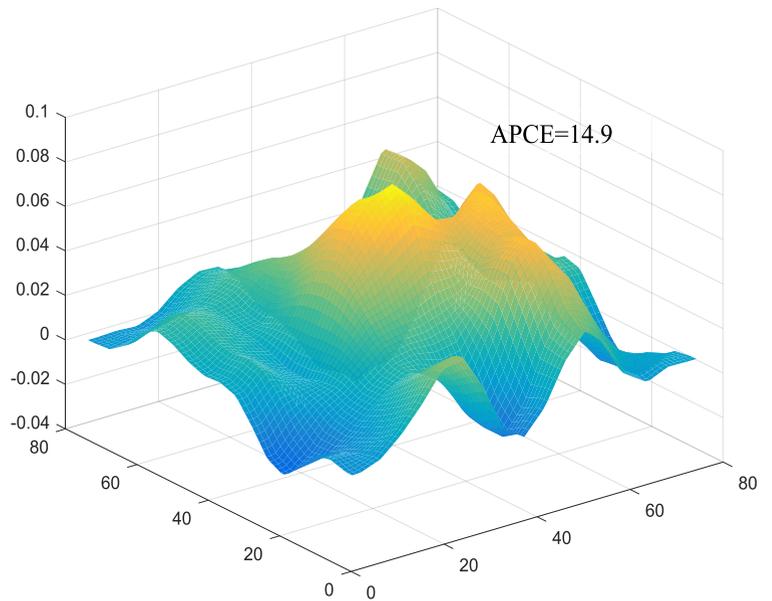

(d)

**Fig. 1. Example of severely deformed or blocked target and correlation filter corresponding graph: (a) original target; (b) correlation filter of the corresponding graph (a); (c) deformed or blocked target; (d) correlation filter of corresponding graph (c). When the target is severely deformed or**

blocked, the confidence response graph presents multi-peak irregularities, and the APCE of the response graph decreases rapidly from 14.9 to 3.6.

In this paper, a concept of relative confidence is utilized for the adaptive adjustment of fusion coefficients. When a target is obstructed, the APCE is decreased. The model is only updated when the APCE exceeds the corresponding average of historical values. The new concept of relative confidence level is used based on APCE to adjust the fusion parameters. The relative confidence is defined as:

$$r_t = \frac{APCE_t}{\sum_{i=1}^{t} APCE_i} , \qquad (23)$$

where $r_t$ is the relative confidence of the detection result at frame $t$. The fusion coefficient $\alpha$ can be adjusted as

$$\alpha_t = \frac{2\alpha}{\left(1+e^{\rho(1-r_t)}\right)} , \qquad (24)$$

where $\alpha_t$ is the weighting coefficient of frame $t$, and $\rho$ is the influencing factor of the relative confidence. The smaller the influence factor is $\rho$, the smaller the influence of relative confidence on the fusion coefficient. In this paper, we set $\rho=1$, $\alpha=0.25$. When a target is obstructed, both APCE and $r_t$ are decreased. When the relative confidence level of the correlation filter is higher than 1, it is more inclined to the detection result of the correlation filter. Conversely, it is more inclined to the detection result of the Bayes classifier.

## D. RECOMMENDATIONS TARGET TRACKING ALGORITHM USING ADAPTIVE FEATURE FUSION

By utilizing the convolution reduction approach (Section C) in conjunction with the adaptive feature fusion between the HOG and GRAY features (Section D), we construct a new target tracking method. Fig.2 shows the overall structure of the proposed tracking algorithm. The object position is achieved by the maximum response value of HOG and GRAY features. Bayesian classifiers' target position is achieved by input target the maximum response value of GRAY features. By adaptive fusion, the target position can be estimated and high confidence model updating and detailed measuring methods are used to improve the stability and accuracy of the tracking algorithm, to prevent interference of drifting model or changing size.

The complete target tracking algorithm includes initialization and updating steps. The input of the proposed target tracking algorithm is the target information of the *t* frame of a video sequence. The output is the target information of the *t*+1 frame of the video sequence. In the initialization step, the extraction of HOG and gray features of a given target is performed. Then, the Kernel correlation filter model is trained by HOG and gray level features, and the Bayesian classifier model is trained by gray level features. The updating step performs six sub-steps. First, HOG and gray features are extracted from the input image, and the feature image is dimensionally reduced and then input to the kernel correlation filter. Second, gray features are extracted from the input image and input to the Bayesian classifier. Third, response diagrams of kernel correlation filter and the Bayesian classifier are obtained respectively. Fourth, the adaptive fusion response map with relative confidence is performed. Fifth, the estimation of the target location for the maximum response value is performed. Sixth, the classifier model with new target information is updated.

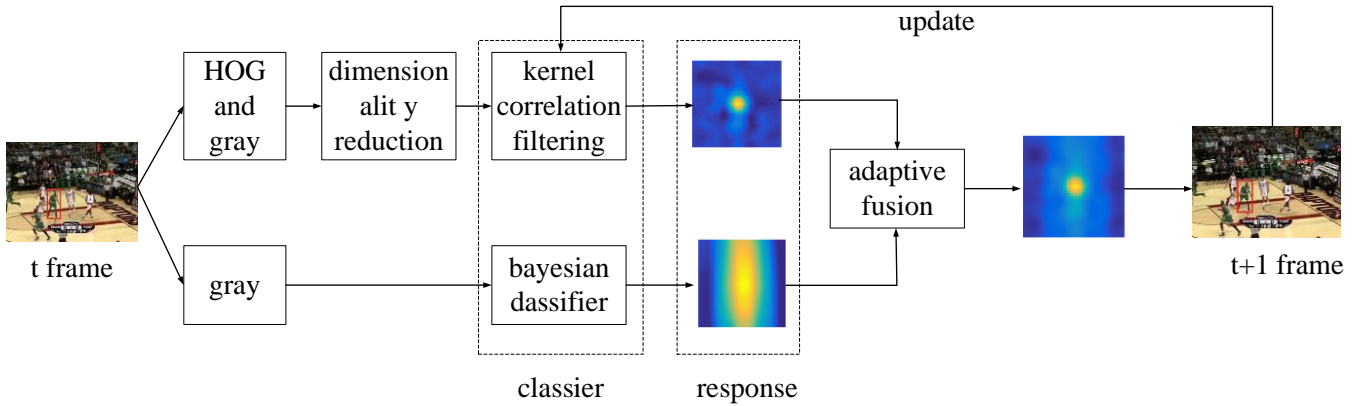

**Fig.2 The overall structure of the proposed tracking algorithm shown in Algorithm 1**

## III. RESULTS

### A. EXPERIMENTAL SETUP

To demonstrate the efficiency and performance of our algorithm, an OTB 100 test set [43] is used in the following experiments. To compare with the proposed method, the 9 target-tracking algorithms, i.e. SRDCF, Staple, fDSST, LCT, DCFNet, KCF, SAMF, DCF, ASLA, are selected. These algorithms are implemented using a MATLAB 2016 platform on a high-perform computer with an intel-i5-4590 CPU, the main frequency of 3.3 GHz, and memory of 8G. In the following experiments, the correlation filter of the proposed method is set as $\lambda=0.001$. The learning rate is $\eta=0.01$, and the feature used is the HOG and gray. For the Bayesian classifier, the learning rate is used as $\theta=0.04$ and the number of histograms is 32.

To quantitatively evaluate the performance of the target tracking algorithms used in the following experiments, an accuracy curve (AC) and a success-rate factor curve (SRFC) are used as the evaluation criteria. The AC is the ratio between the filtered frames' number and the total frame number. In the AC, the center position error (CPE) is the distance between the target position (obtained by the algorithm) and the center position of the target manually marked of filtered frames in the training set. The CPE of filtered frames should be less than 20 pixels. The SRFC is also the ratio between the filtered frame number and the total frame number, where the filtered frames should have an overlap rate of 50%. Here, the overlap rate is the ratio of the overlap area (from two boxes obtained by the algorithm and manually marked by the user) and the total area. The SRFC uses an area size surrounded by the success-rate factor curve to rank the order of nine tracking algorithms.

### B. RESULTS AND DISCUSSION

First, we use the 99 frames of the OTB100 test set to analyze the overall accuracy and success rate of the aforementioned nine algorithms. Figure 2 shows the experimental results. From the experimental results, we found that the accuracy of the proposed algorithm is 0.807, which means it can improve by 2.3% compared to the original STAPLE algorithm. Accordingly, the success rate of the proposed algorithm is 0.598, increasing 1.9% compared with the original STAPLE algorithm. Compared with the second-ranked SRDCF method, the proposed algorithm can improve by 1.9%. Besides, the processing speed of SRDCF is only 8fps, while the proposed algorithm can achieve 50+ FPS, which indicates our method can achieve real-time tracking.

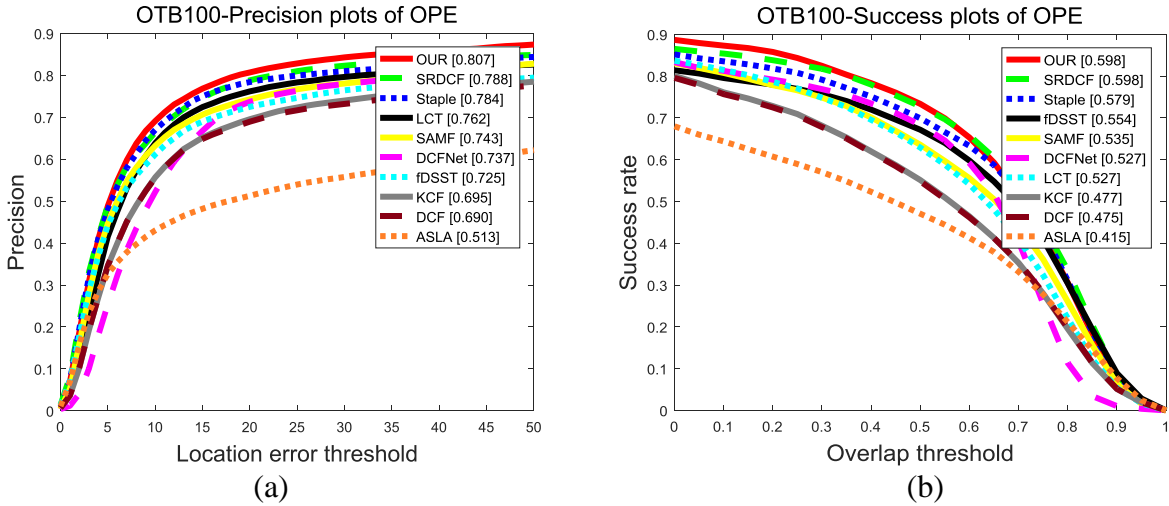

**Fig. 2.** Accuracy and success rate of 10 algorithms under OPE: (a) accuracy (b) success rate.

Second, we also comprehensively analyze the tracking effect of the proposed algorithm under different circumstances. The AC and SRFC of the 10 algorithms are tested using 11 different video attributes. The 11 video attributes include: illumination variation(IV), out of plane rotation(OPR), scale variation(SV), occlusion(OCC), deformation(DEF), motion blur(MB), fast motion(FM), in-plane rotation(IPR), out of view(OFV), background clutter(BC), and low resolution(LR). Figures 3 and 4 reflect the scores of the different algorithms under different video attributes. To intuitively observe the results, the key data from Fig.3 and 4 are demonstrated by Tab.1 and Tab.2. Based on the experimental results, we can observe that the accuracy of the algorithm in this paper ranks first under 8 attributes. Under the other three attributes, the proposed algorithm also ranks second. These results indicate that the algorithm can achieve accurate tracking of the target. In terms of success rate, the algorithm in this paper ranks the first among 7 attributes, and the other 4 attributes also rank the second, indicating that the algorithm in this paper can well solve the problem of target tracking failure in target occlusion, scale change, deformation, and other complex scenarios.

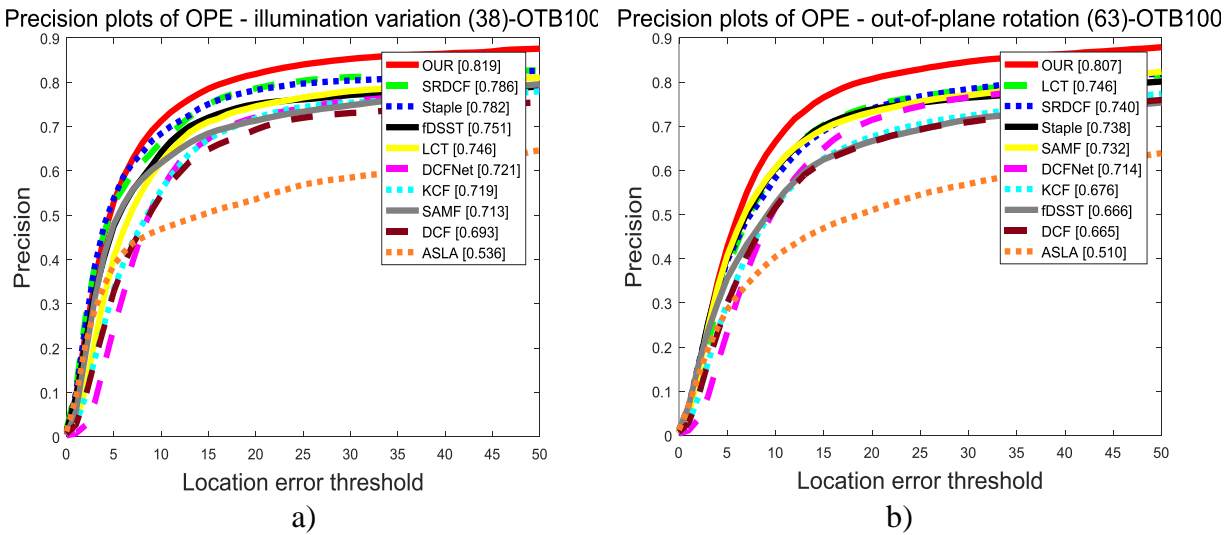

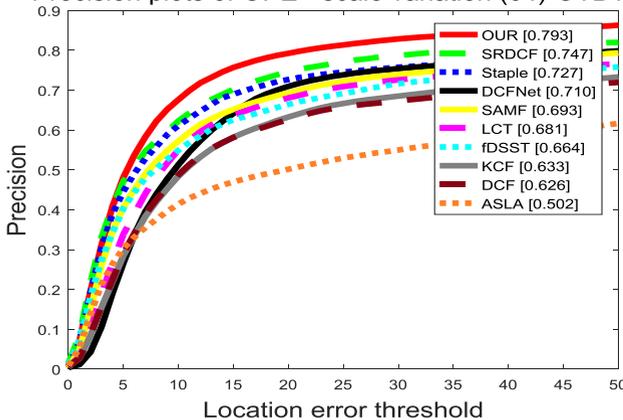

c)

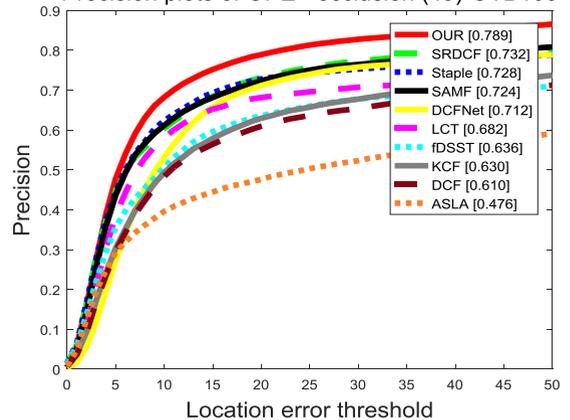

d)

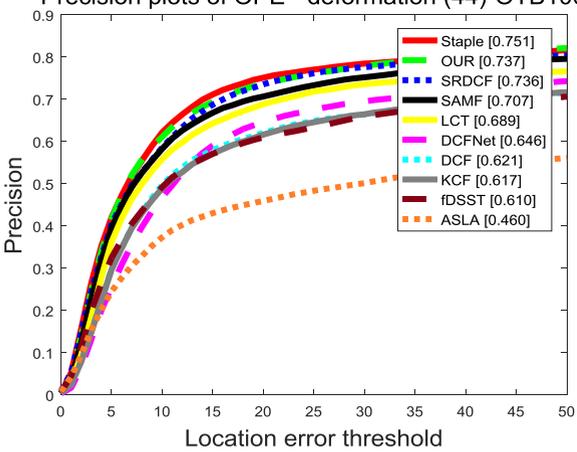

e)

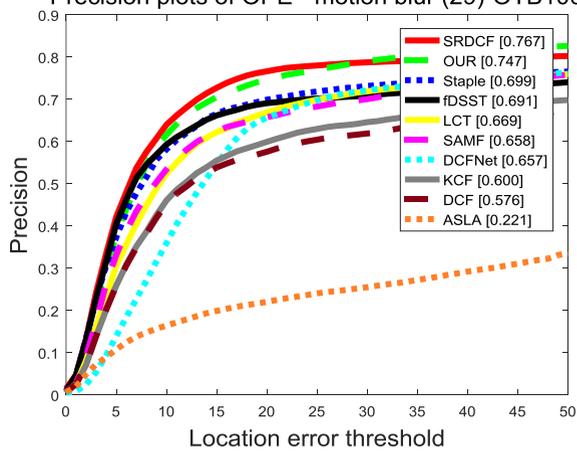

f)

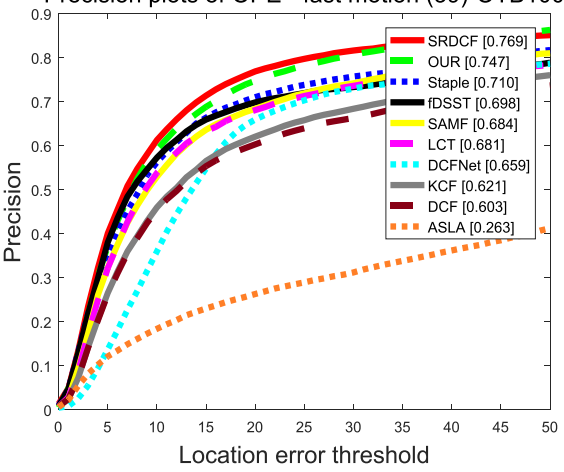

g)

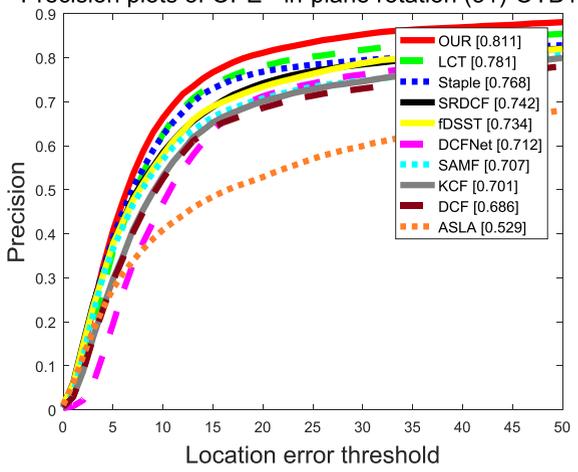

h)

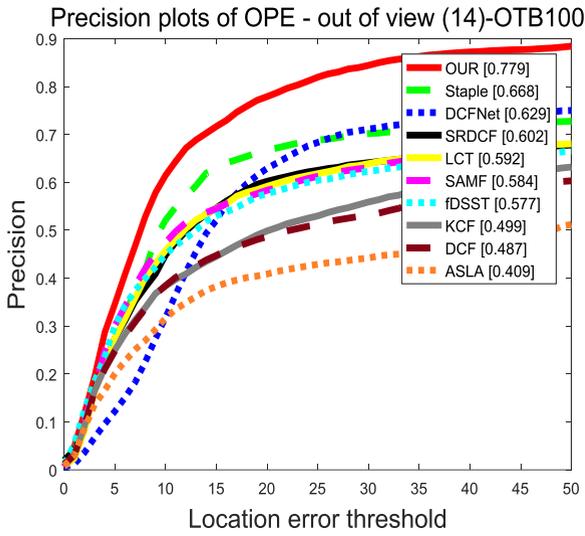

i)

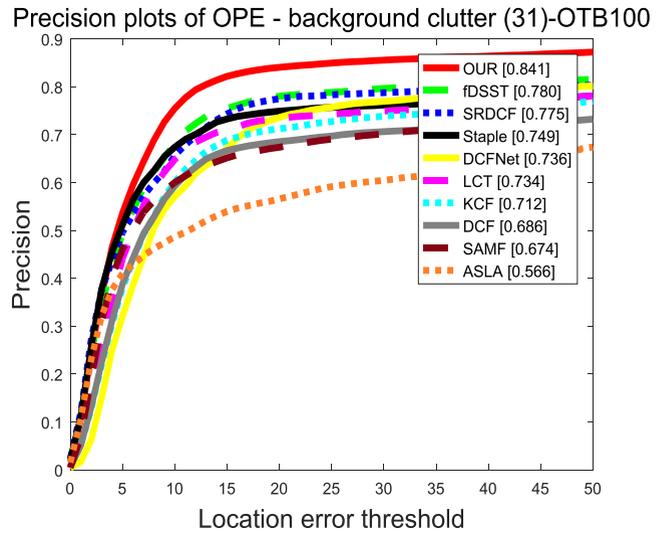

j)

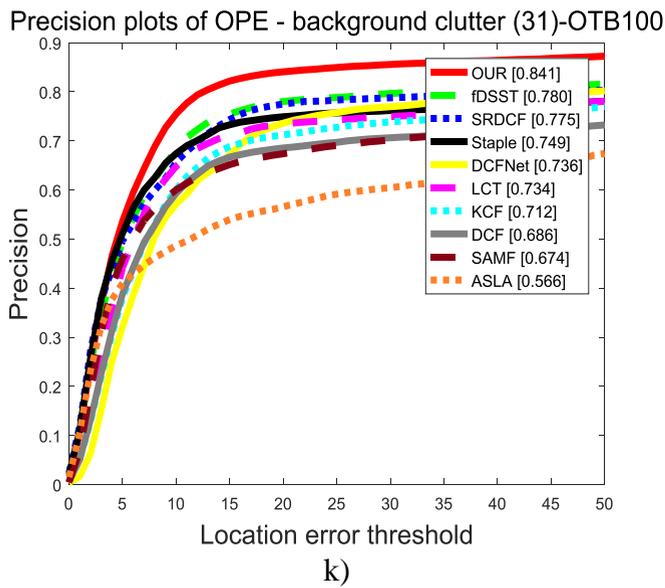

k)

**Fig.3. Accuracy diagrams of 10 algorithms on 11 attributes: (a) illumination variation(IV), (b) out of plane rotation(OPR), (c) scale variation(SV) ,(d) occlusion(OCC), (e) deformation(DEF), (f) motion blur(MB) ,(g) fast motion(FM), (h) in-plane rotation(IPR) ,(i) out of view(OFV), (j) background clutter(BC), (k) low resolution(LR)**

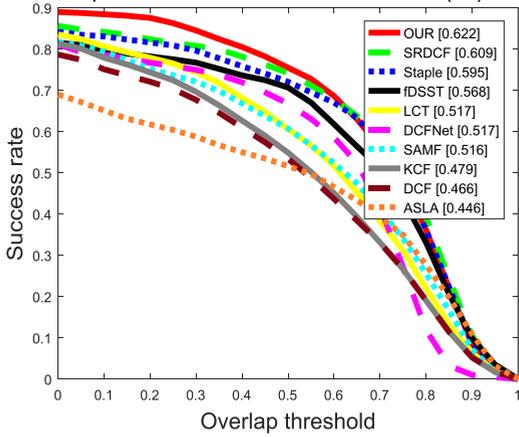
a)
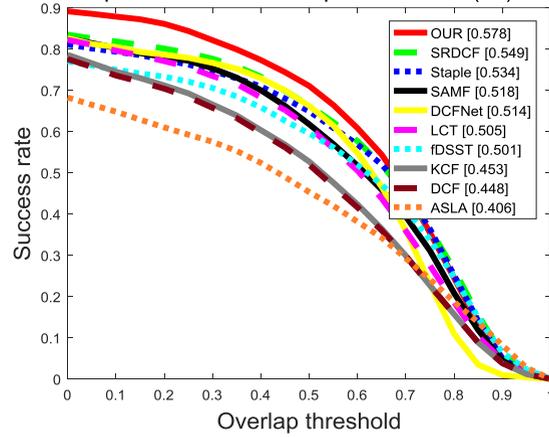
b)
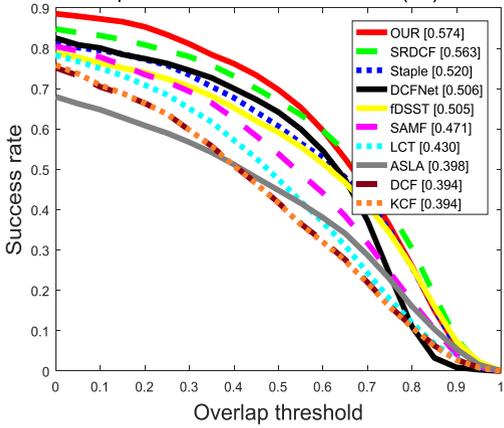
c)
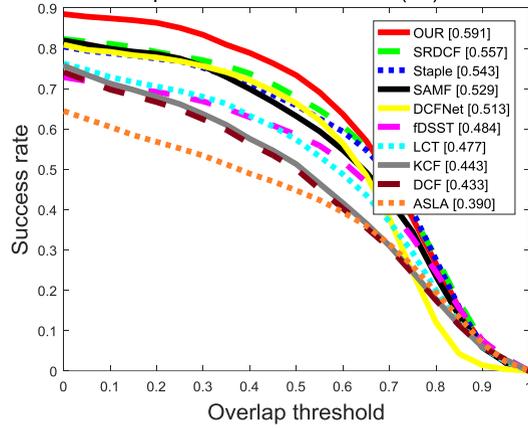
d)
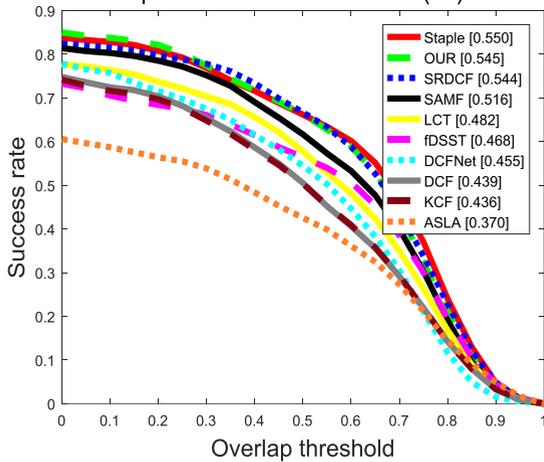
e)
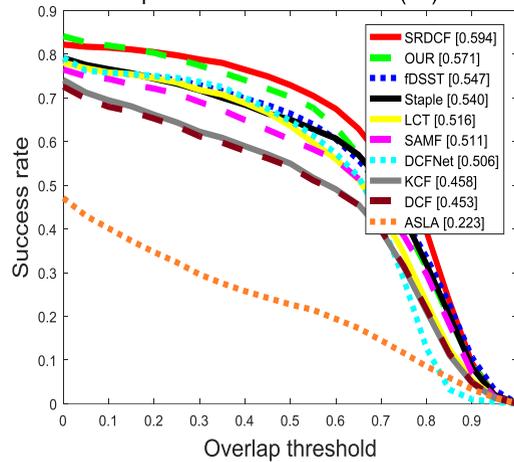
f)

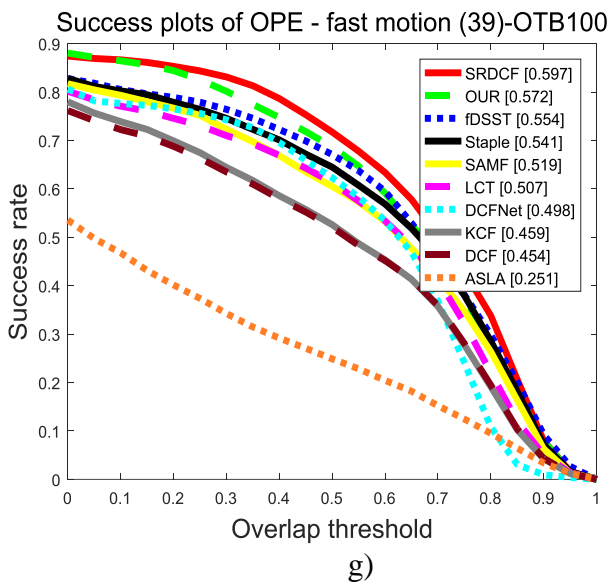

g)

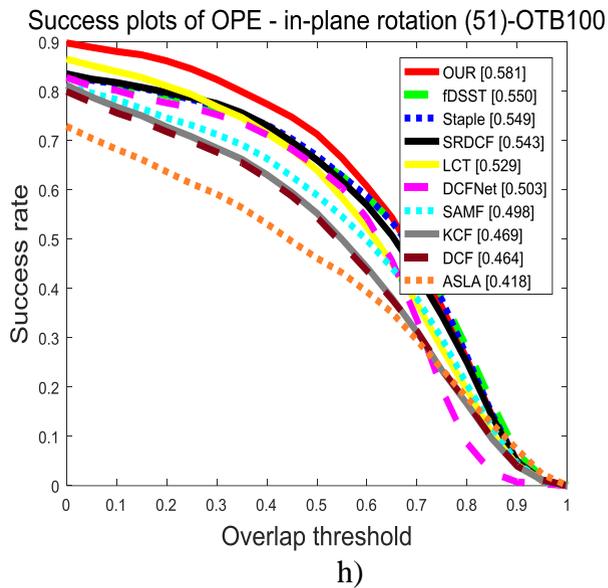

h)

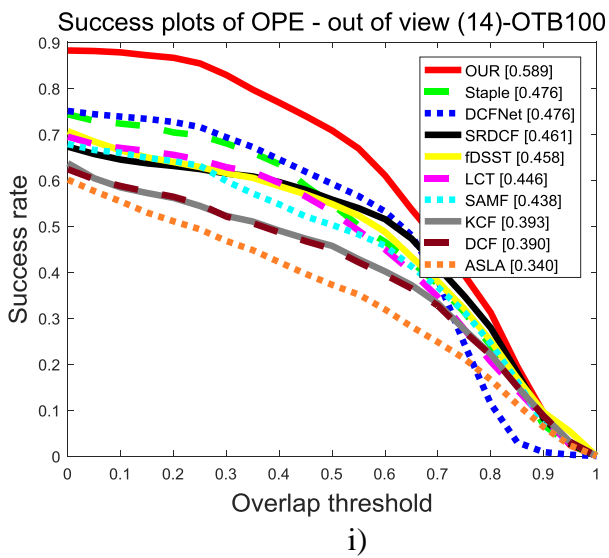

i)

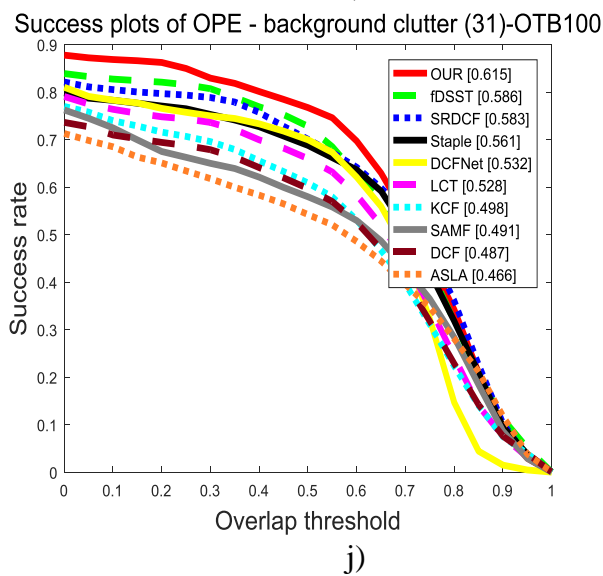

j)

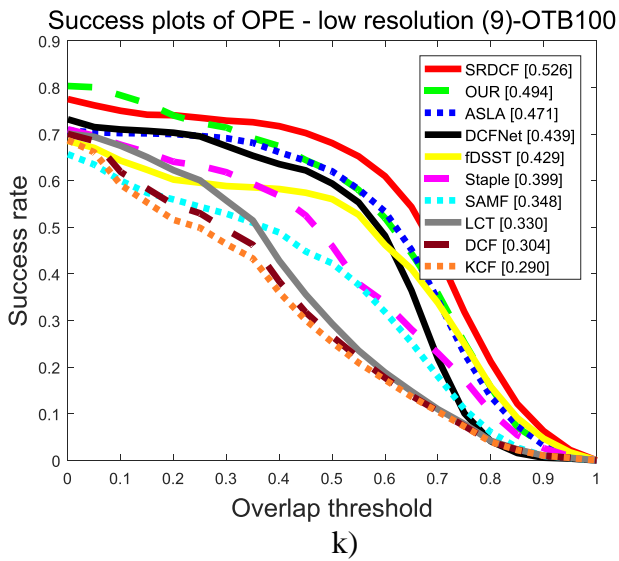

k)

**Fig. 4. Success rate diagrams of 10 algorithms on 11 attributes: (a) illumination variation(IV), (b) out of plane rotation(OPR), (c) scale variation(SV), (d) occlusion(OCC) ,(e) deformation(DEF), (f) motion blur(MB) ,(g) fast motion(FM), (h) in-plane rotation(IPR) ,(i) out of view(OFV), (j) background clutters (BC) ,(k) low resolution(LR)**

TABLE I

ACCURACY TABLES OF 10 ALGORITHMS ON 11 ATTRIBUTES (RED IS THE FIRST OF TEN ALGORITHMS UNDER A CERTAIN ATTRIBUTE, BLUE IS SECOND, GREEN IS THIRD

|        | IV   | OPR  | SV   | OCC  | DEF  | MB   | FM   | IPR  | OV   | BC   | LR   |
|--------|------|------|------|------|------|------|------|------|------|------|------|
| OUR    | 81.9 | 80.7 | 79.3 | 78.9 | 73.7 | 74.7 | 74.7 | 81.1 | 77.9 | 84.1 | 80.0 |
| SRDCF  | 78.6 | 74.0 | 74.7 | 73.2 | 73.6 | 76.7 | 76.9 | 74.2 | 60.2 | 77.5 | 77.4 |
| Staple | 78.2 | 73.8 | 72.7 | 72.8 | 75.1 | 69.9 | 71.0 | 76.8 | 66.8 | 74.9 | 69.5 |
| fDSST  | 75.1 | 66.6 | 66.4 | 63.6 | 61.0 | 69.1 | 69.8 | 73.4 | 57.7 | 78.0 | 67.5 |
| LCT    | 74.6 | 74.6 | 68.1 | 68.2 | 68.9 | 66.9 | 68.1 | 78.1 | 59.2 | 73.4 | 69.9 |
| DCFNet | 72.1 | 71.4 | 71.0 | 71.4 | 64.6 | 65.7 | 65.9 | 71.2 | 62.9 | 73.6 | 71.3 |
| KCF    | 71.9 | 67.6 | 63.3 | 63.0 | 61.7 | 60.0 | 62.1 | 70.1 | 49.9 | 71.2 | 67.1 |
| SAMF   | 71.3 | 73.2 | 69.3 | 72.4 | 70.7 | 65.8 | 68.4 | 70.7 | 58.4 | 67.4 | 63.2 |
| DCF    | 69.3 | 66.5 | 62.6 | 61.0 | 62.1 | 57.6 | 60.3 | 68.6 | 48.7 | 68.6 | 69.4 |
| ASLA   | 53.6 | 51.0 | 50.2 | 47.6 | 46.0 | 22.1 | 26.3 | 52.9 | 40.9 | 56.6 | 70.8 |

TABLE II

SUCCESS TABLES OF 10 ALGORITHMS ON 11 ATTRIBUTES (RED IS THE FIRST OF TEN ALGORITHMS UNDER A CERTAIN ATTRIBUTE, BLUE IS SECOND, GREEN IS THIRD

|        | IV   | OPR  | SV   | OCC  | DEF  | MB   | FM   | IPR  | OV   | BC   | LR   |
|--------|------|------|------|------|------|------|------|------|------|------|------|
| OUR    | 62.2 | 57.8 | 57.4 | 59.1 | 54.5 | 57.1 | 57.2 | 58.1 | 58.9 | 61.5 | 49.4 |
| SRDCF  | 60.9 | 54.9 | 56.3 | 55.7 | 54.4 | 59.4 | 59.4 | 54.3 | 46.1 | 58.3 | 52.6 |
| Staple | 59.5 | 53.4 | 52.0 | 54.3 | 55.0 | 54.0 | 54.1 | 54.9 | 47.6 | 56.1 | 39.9 |
| fDSST  | 56.8 | 50.1 | 50.5 | 48.4 | 46.8 | 54.7 | 55.4 | 55.0 | 45.8 | 58.6 | 42.9 |
| LCT    | 51.7 | 50.5 | 43.0 | 47.7 | 48.2 | 51.6 | 50.7 | 52.9 | 44.6 | 52.8 | 33.0 |
| DCFNet | 51.7 | 51.4 | 50.6 | 51.3 | 45.5 | 50.6 | 49.8 | 50.3 | 47.6 | 53.2 | 43.9 |
| KCF    | 47.9 | 45.3 | 39.4 | 44.3 | 43.6 | 45.8 | 45.9 | 46.9 | 39.3 | 49.8 | 29.0 |
| SAMF   | 51.6 | 51.8 | 47.1 | 52.9 | 51.6 | 51.1 | 51.9 | 49.8 | 43.8 | 49.1 | 34.8 |
| DCF    | 46.6 | 44.8 | 39.4 | 43.3 | 43.9 | 45.3 | 45.4 | 46.4 | 39.0 | 48.7 | 30.4 |
| ASLA   | 44.6 | 40.6 | 39.8 | 39.0 | 37.0 | 22.3 | 25.1 | 41.8 | 34.0 | 46.6 | 47.1 |

Finally, we select four key-frames to demonstrate the tracking results using the STAPLE, KCF, and our algorithm. Fig.5 shows the tracking results, where the red box represents the results of our algorithm. The green box represents the results of the STAPLE algorithm. The blue box represents the results of the KCF algorithm. In Fig.5 (a), the video has plane rotation, motion blur, and fast motion. In Fig.5 (b), the video has a dense covering and similar targets. In Fig.5 (c), the video has occlusion, similar targets, and low resolution. In Fig.5 (d), the video has illumination variation and occlusion. From the experimental results, our algorithm can accurately track the target under Fig.5 (a), (c), and (d). We also found that three algorithms have the tracking shift under Fig.5 (b) because the video includes dense covering and similar targets.

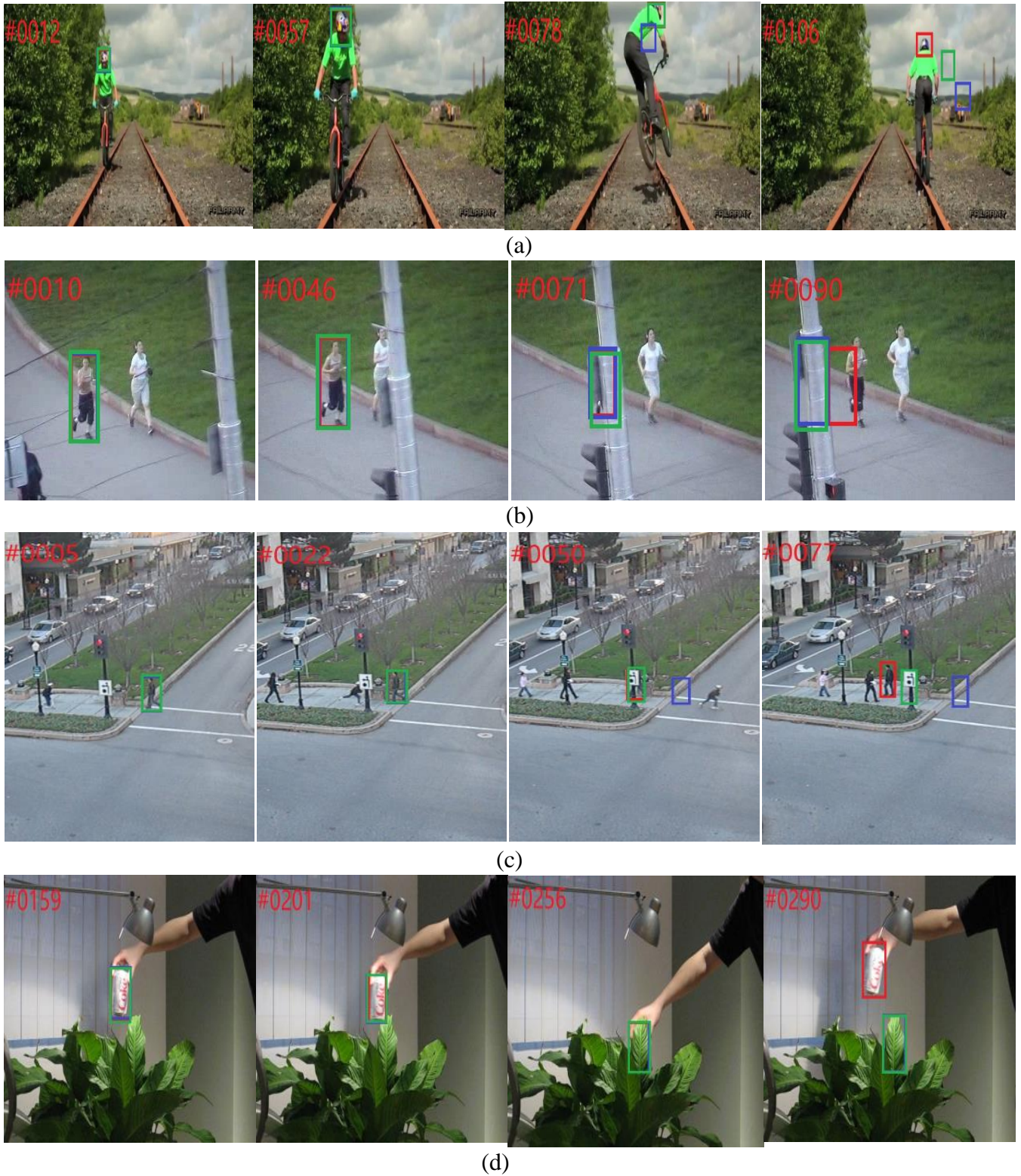

**Fig. 5 The algorithm of this paper (red), STAPLE (green), KCF (blue) in the key sequence tracking graph: (a)Biker, (b) Jogging, (c) Human3, (d) Coke two methods' comparison**

## IV. CONCLUSION

This paper proposes an efficient target tracking algorithm, capable of simultaneous implementation of yielding high tracking accuracy, improving the speed (i.e., real-time) of the algorithm, and preventing the over-fitting issue. The proposed tracking algorithm adopted three advanced methods. First, the adaptive feature fusion between the HOG features and color features is able to improve tracking accuracy. Second, a convolution dimension reduction approach of extracting features is applied to the adaptive fusion to improve the tracking speed. Third, an average correlation energy estimation method is adopted to extract the relative confidence adaptive coefficients to guarantee tracking accuracy. The experiment results indicated that the accuracy and success rate of the algorithm in this paper are 0.807 and 0.598 respectively, which are far higher than the standard data set testing. We provided a more efficient method to capture the target smoothly under different complex surroundings. The proposed method can better solve the problem of target occlusion and scale variation, meeting the requirements of real-time performance and high accuracy. This work has also a potential application to extend the field of 3D holographic video systems [44].

**Funding:** Yanyan Liu would like to thank the Innovation Foundation of Changchun University of Science and Technology (XJJLG-2018-07) and and the Education Department of Jilin Province, China (JJKH20210838KJ) . J. L. is grateful for financial support through a Marie Sklodowska Curie Fellowship, H2020-MSCA-IF-2020-101022219.

**Conflict of Interest:** Authors declares that they have no conflict of interest.

**Ethical approval:** This article does not contain any studies with human participants or animals performed by any of the authors.

# REFERENCES

[1] Ying Li, Pengchen Li, Qiang shen, Real-time infrared target tracking based on $\ell 1$ minimization and compressive features, Appl. Opt. 53(28) (2014) 6518-6526.
[2] A. Bal, M.S. Alam, Dynamic target tracking with fringe-adjusted joint transform correlation and template matching, Appl. Opt. 43 (25) (2004) 4874–4881.
[3] B.-S. Jeung, S.-S. Lim, D.-H. Lee, Development of optical sighting system for moving target tracking, Current Optics and Photonics 3 (2) (2019) 154–163.
[4] Zhou Yi, Zhou Sheng Tong, Zhong Zuo Yang, Li Hong Guang, A de-illumination scheme for face recognition based on fast decomposition and detail feature fusion, Opt Express 21(9) (2013) 11294-11308.
[5] D.S. Bolme, J.R. Beveridge, B.A. Draper, Y.M. LuiVisual, Object tracking using adaptive correlation filters, Computer Vision & Pattern Recognition. 119 (5) (2010) 2544–2550.
[6] Wang, Xinchao, et al., Tracking interacting objects using intertwined flows, IEEE transactions on pattern analysis and machine intelligence, 38(11) (2015) 2312- 2326.
[7] E. Türetken, et al., Network flow integer programming to track elliptical cells in time-lapse sequences, IEEE Trans. Med. Imaging 36 (4) (2016) 942–951.
[8] X. Wang, et al., Greedy batch-based minimum-cost flows for tracking multiple objects, IEEE Trans. Image Process. 26 (10) (2017) 4765–4776.
[9] J.F. Henriques, C. Rui, P. Martins, et al., Exploiting the Circulant Structure of Tracking-by-Detection with Kernels, in: European Conference on Computer Vision. Springer-Verlag, 2012, pp. 702-715.

[10] J.F. Henriques, C. Rui, P. Martins, J. Batista, High-speed tracking with kernelized correlation filters, IEEE Trans. Pattern Anal. Mach. Intell. 37 (3) (2015) 583–596.
[11] M. Danelljan, G. Häger, F.S. Khan, M. Felsberg, Accurate Scale Estimation for Robust Visual Tracking, in: British Machine Vision Conference, BMVC, 2014.
[12] L. Bertinetto, J. Valmadre, S. Golodetz, O. Miksik, P.H.S. Torr, Complementary learners for real-time tracking, Computer Vision & Pattern Recognition 38 (2) (2016) 1401–1409.
[13] M. Danelljan, F.S. Khan, M. Felsberg, J. van de Weijer, Adaptive Color Attributes for Real-Time Visual Tracking, In CVPR, 2014.
[14] G Liu, S Liu, K Muhammad, AK Sangaiah, Faiyaz, Object tracking in vary lighting conditions for fog based intelligent surveillance of public spaces, IEEE Access., vol.6, Jun. 2018, pp.29283-29296.
[15] N. Dalal, B. Triggs, Histograms of oriented gradients for human detection, in: Proc. IEEE Comput. Soc. Conf. Comput. Vis. Pattern Recog- nit. (CVPR), vol. 1, Jun. 2005, pp. 886–893.
[16] Y. Wang, Q. Pan, C. Zhao, Y. Cheng, Fuzzy color histogram based kernel tracking under varying illumination, in: Proc. 4th Int. Conf. Fuzzy Syst. Knowl. Discovery (FSKD), vol. 4, Aug. 2007, pp. 235–239.
[17] A. Akhondi-Asl, S.K. Warfield, Estimation of the prior distribu- tion of ground truth in the staple algorithm: an empirical Bayesian approach, Proc. MICCAI (2012) 593–600.
[18] S.K. Warfield, K.H. Zou, W.M. Wells, Simultaneous truth and performance level estimation (STAPLE): an algorithm for the valida- tion of image segmentation, IEEE Trans. Med. Imag. 23 (7) (Jul. 2004) 903–921.
[19] L. Bertinetto, J. Valmadre, S. Golodetz, O. Miksik, Philip H.S. Torr, Staple: Complementary Learners for Real-Time Tracking, in: IEEE Conference on Computer Vision and Pattern Recognition (CVPR), 2016, pp. 1401-1409.
[20] A. Akhondi-Asl, L. Hoyte, M.E. Lockhart, S.K. Warfield, A Logarithmic Opinion Pool Based STAPLE Algorithm for the Fusion of Segmentations With Associated Reliability Weights, IEEE Trans. Med. Imag. 33 (10) (Oct. 2014) 1997–2007.
[21] J. Valmadre, L. Bertinetto, F.S. Khan, M. Felsberg, End-to- end representation learning for correlation filter based tracking.'' [Online], 2017, Available: htt ps://arxiv.org/abs/1704.06036.
[22] Error-space estimation method and simplified algorithm for space target tracking, Applied Opt. 49(38) (2010) 5384-5490.
[23] K.S. Kaawaase, F. Chi, J. Shuhong, Q. Bo Ji, A review on selected target tracking algorithms, Information Technology Journal, (2011) 691-702.
[24] [24] Z. Wang, Q. Hou, L. Hao, Improved infrared target-tracking algorithm based on mean shift, Applied Opt. 51 (21) (2012) 5051–5059.
[25] B.-S. Jeung, S.-S. Lim, D.-H. Lee, Development of optical sighting system for moving target tracking, Curr. Optics Photonics 3 (2) (2019) 154–163.
[26] S. Savazzi, M. Nicoli, F. Carminati, M. Riva, A Bayesian approach to device-free localization: modeling and experimental assessment, IEEE J. Sel. Top. Signal Process. 8 (1) (Feb. 2014) 16–29.
[27] Cetin, Mujdat, Ivana Stojanovic, Ozben Onhon, Kush Varshney, Sadegh Samadi, William Clem Karl, Alan S. Willsky, Sparsity-Driven Synthetic Aperture Radar Imaging: Reconstruction, autofocusing, moving targets, and compressed sensing, IEEE Signal Processing Magazine 31(4) (July 2014) 27–40.
[28] St´ephanie Bidon, Jean-Yves Tourneret, Laurent Savy, Fran çois Le Chevalier, Bayesian sparse estimation of migrating targets for wideband radar, IEEE Transactions on Aerospace and Electronic Systems, 50(2) (July 2014) 871-886.

[29] L. Mihaylova, A.Y. Carmi, F. Septier, A. Gning, S. Pang, S. Godsill, Overview of Bayesian sequential Monte Carlo methods for group and extended object tracking, Digital Signal Process. 25 (February 2014) 1–16.
[30] S. Weerakkody, B.Sinopoli, Detecting integrity attacks on control systems using a moving target approach, in: IEEE Annual Conference on Decision and Control, Dec. 2015, pp. 5820-5826.
[31] S. Buzzi, M. Lops, L. Venturino, M. Ferri, Track-before-detect procedures in a multi-target environment, IEEE Trans. Aerosp. Electron. Syst. 44 (3) (July 2008) 1135–1150.
[32] L. Yan, Yin Lu, Yerong Zhang, An Improved NLOS Identification and Mitigation Approach for Target Tracking in Wireless Sensor Networks, IEEE Access, 5 (March 2017) 2798-2807.
[33] G. Newstadt, E. Zelnio, A. Hero, Moving target inference with bayesian models in SAR imagery, IEEE Trans. Aerosp. Electron. Syst. 50 (3) (July 2014) 2004–2018.
[34] J. Han, K.-K. Ma, Fuzzy color histogram and its use in color image retrieval, IEEE Trans. Image Process. 11 (8) (Aug. 2002) 944–952.
[35] JiFeng Ning, L. Zhang, D. Zhang, W.u. ChengKe, Robust Object Tracking Using Joint Color-trxture Histogram, Int. J. Pattern Recognit. Artif. Intell. 23 (7) (Aug. 2009) 1245–1263.
[36] Xiaolin Wu, Bin Zhu, Yutong Hu, Yamei Ran, A Novel Color Image Encryption Scheme Using Rectangular Transform-Enhanced Chaotic Tent Maps, IEEE Access., 5 (May 2017) 6429–6436.
[37] P. Praveenkumar, R. Amirtharajan, K. Thenmozhi, J.B.B. Rayap-pan, 'Triple chaotic image scrambling on rgb–a random image encryp- tion approach', Secur. Commun. Netw. 8 (18) (2015) 3335–3345.
[38] F.-G. Jeng, W.-L. Huang, T.-H. Chen, 'Cryptanalysis and improve- ment of two hyper-chaos-based image encryption schemes', Signal Process. Image Commun. 34 (May 2015) 45–51.
[39] A. Adam, E. Rivlin, I. Shimshoni, Robust Fragments-based Tracking using the Integral Histogram, in: IEEE Computer Society Conference on Computer Vision and Pattern Recognition(CVPR), June 2006, pp. 17–22.
[40] J. Hanfner, H.S. Sawhney, W. Equitz, M. Flickner, W. Niblack, Efficient color histogram indexing for quadratic form distance functions, IEEE Trans. Pattern Anal. Mach. Intell. 7 (7) (Jul. 1995) 729–736.
[41] Rishav Chakravarti, Xiannong Meng, A Study of Color Histogram Based Image Retrieval, in: International Conference on Information Technology: New, Jul. 2009, pp. 1323–1328.
[42] D.S. Bolme, J.R. Beveridge, B.A. Draper, Y.M. LuiVisual, Object Tracking using Adaptive Correlation Filters, Computer Vision & Pattern Recognition (2010) 2544–2550.
[43] Y. Wu, J. Lim, M.-H. Yang, Online object tracking: A benchmark, IEEE Computer vision and pattern recognition (CVPR), 2013, pp. 2411-2418.
[44] J. Li, Q. Smithwick, D. Chu, Holobricks: modular coarse integral holographic displays, Light: Science and Applications 11 (1) (2022) 1–15.